# Computer-Aided Modelling of the Bilingual Word Indices to the Ninth-Century Uchitel'noe evangelie

Martin Ruskov[1], Lora Taseva[2]

[1]*University of Milan, 20100 Milan, Italy*

[2]*Institute of Balkan Studies and Centre of Thracology, Bulgarian Academy of Sciences, 1000 Sofia, Bulgaria*

**Abstract**
The development of bilingual dictionaries to medieval translations presents diverse difficulties. These result from two types of philological circumstances: a) the asymmetry between the source language and the target language; and b) the varying available sources of both the original and translated texts. In particular, the full critical edition of Tihova of Constantine of Preslav's *Uchitel'noe evangelie* ('Didactic Gospel') gives a relatively good idea of the Old Church Slavonic translation but not of its Greek source text. This is due to the fact that Cramer's edition of the catenae – used as the parallel text in it – is based on several codices whose text does not fully coincide with the Slavonic. This leads to the addition of the newly-discovered parallels from Byzantine manuscripts and John Chrysostom's homilies. Our approach to these issues is a step-wise process with two main goals: a) to facilitate the philological annotation of input data and b) to consider the manifestations of the mentioned challenges, first, separately in order to simplify their resolution, and, then, in their combination. We demonstrate how we model various types of asymmetric translation correlates and the variability resulting from the pluralism of sources. We also demonstrate how all these constructions are being modelled and processed into the final indices. Our approach is designed with generalisation in mind and is intended to be applicable also for other translations from Greek into Old Church Slavonic.

## 1. Introduction

The creation of the Slavonic alphabet in the second half of the 9th century marked the beginning of a completely new written tradition in Europe – first only in Glagolitic script, then in Cyrillic. The main part of this literary production consisted of translated texts from Greek, which were linked to the needs of the church practice. The studies on the relationship between source and target texts have occupied a significant place in the philological Slavic Medieval Studies (Palaeoslavistics) and an important tool for their analysis have been the bilingual dictionaries to specific scholarly edited writings. These diachronic dictionaries provide the scholars not only with a generalised view to the vocabulary of the respective work but also with an opportunity to study various specific issues concerning the Greek-Slavonic translation correlates.

The very process of developing such dictionaries is complicated not just from the philological perspective but also from the perspective of their computer support. A major technical challenge in this respect is the asymmetry between the two languages, i.e., the cases in which there is no *one-to-one* correlation of the words of the original and their translations. We present the approach we have developed for computer-aided creation of such dictionaries, together with the main problems we have encountered and the decisions we have made to overcome them. This paper is structured by the following sections: philological basis, technical approach, overview of specific problems and their solutions, and conclusion.

## 2. Philological basis

The philological background of every bilingual dictionary to a translated text is made up of the two respective sources – original and target ones. In most cases the medieval translations are not preserved as authentic documents, so scholars build their understanding of them on the basis of the preserved (or available) today copies and/or editions. Therefore, the text-critical tradition of the original text and its translation, as well as the degree of the research on them, are factors which directly affect both the content and appearance of the end lexicographic product. Here follows a brief overview of the sources in relation to the bilingual word indices to the Old Church Slavonic *Uchitel'noe evangelie* (UE).

### 2.1. Slavonic tradition

UE was created by Constantine of Preslav, a prominent Old Bulgarian man of letters, probably between the years 886 and 893 [1: 3–4, 10; 2: 86]. It consists of 51 sermons commenting on the respective Sunday Gospel readings for the whole year. The main part of this commentaries is translated from Greek, but the prologue to the collection and the most of the introductory and concluding words to each sermon are authored by Constantine of Preslav.

The manuscript tradition of UE is not complicated – 4 full copies came down to us [3]: one Russian – MS ГИМ, Син. 262 (henceforth S) dated to the late 11th [4: LXIX] or mid-12th century [5] and three Serbian ones – MS РНБ, Гильф. 32 (henceforth G) of the year 1286 [6: 214], MS ÖNB, Cod. Slav. 12 (henceforth W) of the 14th century [7: 126–127] and MS Hil. 385 (henceforth H) of the year 1344 [8: 151–152]. None of them preserves the original text of Constantine in its authentic form, yet they correct and add to each other.

This means that in order to select the most probable original translation correspondences in the word indices, the material from all preserved manuscripts needs to be critically used. The diplomatic edition of the oldest witness by Tihova which includes the variant readings after the other three copies [9] together with the corrections published in the review by Krys'ko [10] give a solid basis for such a lexicographic work.

### 2.2. Greek tradition

No Byzantine collection, identical to UE, is known, but even the early researchers Gorsky and Nevostruyev assume that Constantine of Preslav used some kind of an abridgement of John Chrysostom's Gospel homilies [11: 412, 423–424], and Antonij the archimandrite [12: 37, 40–50] points out the presence of parallels in Greek Gospel catenae published by Cramer [13]. Using the same multivolume publication, Tihova adds to her edition the Greek counterparts for the majority of the translated parts in UE. Her omissions (mainly due to the fact that Cramer's *Supplementum* was unavailable to her) are added by Krys'ko [10]. Unfortunately, Cramer's edition is based on a limited source basis for the Greek catenae with main evidence the Parisian manuscript Cod. Coislin. Gr. 23 (concerning the manuscripts used by Cramer see the work of Lamb [14: 281–290]). The project team, therefore, has broadened the range of sources by adding to it, on the one hand, certain Byzantine catenae manuscripts which are available in both the database of the Münster Institute for New Testament Textual Research and the websites of major libraries such as the ones in Paris, Munich and Vatican, and, on the other, the full text of John Chrysostom's homilies [15]. Kotova [16] and Petrov [17] found in such sources more accurate correspondences for individual words and expressions, parallels for hitherto unidentified fragments, including even some parts of sermons 19, 20 and 42 which were previously considered original. These discoveries reveal that the Greek texts to be processed for the sake of the index should include not just the main text in Cramer's edition and in its *Supplementum* but also the additions and the better readings from the manuscript traditions of the catenae and the full Crhysostomian homilies. What is more, the final index is supposed to include the information about the source from which a particular reading was taken.

**2.3. Language and text asymmetry**

The complexity of these dictionaries, imposed by the two textual traditions and the plurality of sources for each, is combined with the usual difficulties brought about by the asymmetry between the two languages. It is the reason that the lexical units from the original and its translation do not always correspond unambiguously, and also that often a given grammatical meaning is conveyed by lexical means or vice versa. UE, just like other earlier Old Church Slavonic translations, is freer than the later ones because the literary language is still in process of creation, which also means that the stable correlates are rare and the possibilities for variations (both qualitative and quantitative) are much greater. This suggests a significant number of cases beyond the standard models which need specific solving.

## 3. Approach

Automatic alignment between original and translated text is a common task in corpus linguistics [18, 19] . It is performed on languages that are rich with linguistic resources and most often it is limited to alignment at the sentence level. Word-level alignment is a difficult task with limited success rates [19]. When it comes to languages with less resources available, especially old and regional ones, automation is not possible and the use of technology is limited to supporting the work of philologists [20, 21, 22] [6, 10, 11]. This is especially true for texts such as UE, where there is a rich variability both in the Greek original (because of the absence of an exact source text), and its Slavonic translation (because the initial translation is judged by the extant later copies).

Our approach is based on the scheme for the production of dictionaries to the Church Slavonic translations of the anti-Latin polemic of Gregory Palamas and Barlaam of Calabria, developed by von Waldenfels and Taseva [21]. However, due to the greater complexity of the indices to the UE, the approach has had to be revised and expanded to meet the philological goal of approximating, as close as possible, between the Greek sources and their translation. The solution presented here uses two particular open technologies: the Python programming language and the Open Office XML file standard. The result of the process is the generation of two word-indices: Slavonic-Greek and Greek-Slavonic. Each of these maps each lexeme from one of the texts to all the word usages corresponding to it in the other language.

| A | B | C | D | E | F | G | H | I | J | K | L | M | N | O | P | Q | R | S |
|---|---|---|---|---|---|---|---|---|---|---|---|---|---|---|---|---|---|---|
| въ | WGH | въ | въ + Loc. | 1/7d1 | оу | оу насъ | оу praep. | оу + Gen. | | | om. | | | | | παρ' | παρά | παρά + Acc. |

**Figure 1:** A sample row from the input table.

This publication examines in detail the last crucial step of a wider process whose initial stages are as follows. It begins with a transcription of the source text (in this case Tihova's edition of UE) carried out by Rabus with the program Transkribus (for the program see [23]). The text document is then corrected and enriched by a philologist who manualy introduces positional variants from the Slavonic transcriptions. Then it is automatically transformed into a spreadsheet table where each line contains a word from the text and its possible variants. The project team then enriches this table, adding Greek correspondences (including variants), lemmatising the word usages (bringing them into standard dictionary form), and adding necessary grammatical and lexicographical annotations. As a result, the completed table (see Fig. 1) contains the positional correspondences between the source and its target, annotated with the necessary philological information. This table is the input solution described here.

The structure of this table should allow for the representation of variant readings, yet be easily maintainable for philologists working in teams. The proposed solution for this trade-off is summarised in Table 1 and detailed here. Since the focal point of this work is the Slavonic translation, the used Slavonic word form (column F) in the main text, its lemma (columns H) and their sublemmas (in case where it is necessary to explain the correspondence in the other language, sequentially in columns I-K) are central. Next to these are their addresses in the text, consisting of word, page, column, and line (column E), and their contexts, containing the entire line from which the corresponding word was extracted (column G). The information in column

**Table 1**
Reference of used table columns.

|  | Slavonic variants | Slavonic main | Greek main | Greek variants |
|---|---|---|---|---|
| word address |  | E |  |  |
| word | A | F | L | Q |
| word context |  | G |  |  |
| lemma | B | H | M | R |
| first sublemma | C | I | N | S |
| second sublemma | D | J | O | T |
| third sublemma |  | K | P |  |

въ
| въ + Loc.
    παρά + Acc. → παρά: въ WGH/παρ' C (1/7d1^{WGH-C} » оу S)
оу praep.
| оу + Gen.
    παρά + Acc. → παρά: оу/παρ' C (1/7d1^C » [въ WGH])

παρά
| παρά + Acc.
    въ + Loc. → въ: παρ' C/въ WGH (1/7d1^{WGH-C} » оу S)
    оу + Gen. → оу praep.: παρ' C/оу (1/7d1^C » [въ WGH])

въ (1^{var})
| въ + Loc. (1^{var})
    • παρά + Acc. → παρά (1^{var}): 1/7d1^{WGH-C} » оу + Gen. S
оу praep. (1)
| оу + Gen. (1)
    • παρά + Acc. → παρά (1^{var}): 1/7d1^C » [въ + Loc.^{WGH}]

παρά (1^{var})
| παρά + Acc. (1^{var})
    • въ + Loc. → въ (1^{var}): 1/7d1^{WGH-C} » оу + Gen. S
    • оу + Gen. → оу praep. (1): 1/7d1^C » [въ + Loc.^{WGH}]

**Figure 2:** Generated outputs corresponding to Fig. 1. Up left is the generated Slavonic-Greek list, up right – Slavonic-Greek index. Accordingly, below are Greek-Slavonic list and index. For simplification the example is considered in isolation. In real cases these will be aggregated with other uses.

G is not used by the software, but is useful for the annotation work of philologists. To the right of the columns for the main Slavic text is the corresponding information for the Greek main text – word usages (column L) and their adjacent lemmas (columns M-P). Where variant information is available, it is added in the free columns on the appropriate side. In the columns to the left of those for the main Slavic text is the information on Slavic variants – word usages (column A) and lemmas (columns B-D) respectively, and to the right of the main Greek text is the information on Greek variants – word usages (column Q) and lemmas (columns R-T). When more than one variant reading is present, the information is combined in the same variant columns, as illustrated in the last example of Section 4.2.

In the last step of the wider process, the two types of dictionaries are created: lists and indices (see Fig. 2). The *lists* contain all the positional correspondences in the two texts united under a common lemma. Entries show specific word usages and a precise address. These lists are used by philologists to verify alignment and lemmatisation, to correct possible inconsistencies and errors in the collaborative enrichment. The *indices* present the final word indices in a form ready for publication. They contain only the summary information about the lexical correspondences, a list of the exact address of the lemma occurrences and information about their frequency in the text. For each of the two types of dictionaries, both Slavonic-Greek and Greek-Slavonic versions are created.

Two dedicated software tools are used to create the lists and indices using shared logic in the software implementation, the main elements of which are described in this section. The lists are generated by a program called *integrator*. The indices are created by a program called *generator*. Each of the two tools performs three steps, executed sequentially for each of the two dictionaries – Slavonic-Greek and Greek-Slavonic. The first two of these steps are common, the third is different for each of the tools:

1. Adaptation
2. Aggregation
3. Export

**Adaptation** of the content extraction table. This step takes into account which direction the dictionary is currently being built for and creates in-memory tables similar to the input

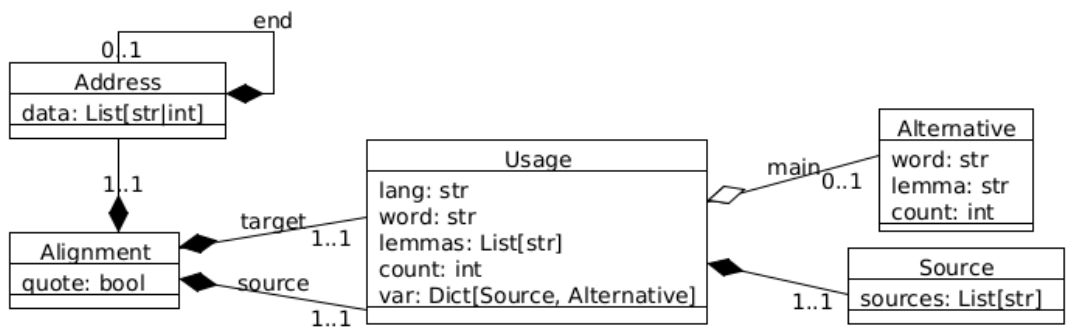

**Figure 3:** UML class diagram for the model of Alignment.

table, but with transformations that allow all the information needed for a single word use to be contained in a single row. This is relevant in cases of quantitative asymmetry (see next section), where this condition is not satisfied for the input table. If there are variants in the information that enters the new row, the step considers separately the columns for the main text and the ones for the variants. This way the necessary adaptive transformation can be applied for each of the two cases separately. This transformation is an application of the *divide-and-conquer* principle. In this case it is important, firstly, for performance reasons. Secondly, it reduces the task of constructing the list or index to the simpler subtask of generating entries from a single row in the adapted table. The simplified subtask is solved in the following step.

**Aggregation** of alphabetically ordered indices. This step is implemented using nested *sorted mapping*[1] standard data structures, which allow the construction of specialised alphabetically sorted reference tables functionally similar to the desired end result. Because this structure is shared by the *integrator* and the *generator*, it must collect the necessary information for both tools to allow the data needed for each to be retrieved and displayed. To make this possible, a grouping is created in the sorted mapping structure according to the following hierarchy:

$$lemma \rightarrow (sublemma \rightarrow (sublemma2 \rightarrow (sublemma3 \rightarrow)))alignment$$

where the parenthesis indicate optional sublemmas, included only if present in the input table.

We call *Alignment* the structure shown in Fig. 3. It consists of the combination of *Usage* (i.e. the word usage) in the source and target languages, an *Address*, and information about whether the text is a Biblical quotation. The Address allows to locate the word usage up to the line (containing also page and column) or up to a line span (represented by the self-reference in address indicating the end of the span in Fig. 3). Biblical quotations are noted in the input tables in bold and italics, and accordingly appear in the same way in the output dictionaries, an example being present in Fig. 7. *Usage* in turn has several important properties. First, these include information about the respective word, its lemmas and, if there are repetitions at this address, a counter specifying the particular repetition of the word in the line, if any. The language of the *Usage* is also stored because of the direction-agnostic approach (see last paragraph of this section), which applies the same logic to the source-to-target and target-to-source dictionaries. Last but not least, the structure stores information about the source (*Source*) of the usage, and possible alternative spellings (*Alternative*), divided into main and a mapping (*var*) of variants to their alternative spellings. This structure contains the necessary information for the needs of the dictionaries created in the next step.

---

[1] In Python called *SortedDict*, available as part of https://pypi.org/project/sortedcontainers/

**Export** of the dictionaries to a word processing document formatted according to their respective needs – for the *integrator* and the *generator*. The result is grouped according to the hierarchy of the previous step. The lemmas are at its highest level. They are sorted alphabetically. They list the usages and possible references to other corresponding variants. In the case of sublemmas at different levels, they divide the usages in the corresponding lemma into separate lines formatted with indentation. Since the *integrator*'s goal is to allow for the verification of the manual preparation of the input table, in addition to the lemmas, the specific word usages are displayed in the generated *lists*. However, for the indices these word usages are superfluous information. Rather than that, the *generator* adds to the *indices* a frequency count of occurrences for each of the languages – both in the main text and in the variants (the latter are indicated with var in superscript). In Fig. 2 the result from the sample input from Fig. 1 is shown. Note that in the example, the word in the Greek main text is missing (denoted by "om.") and, accordingly, in the Greek dictionaries (the respective *list* and *index*), only one lemma appears (παρά). And the two lemmas in the Slavonic dictionaries (въ and оу) are paralleled with only one Greek word from an unspecified variant (παρά).

Our approach applies one more simplifying technique – the Slavonic-Greek indices and the Greek-Slavonic indices are treated as symmetric. In other words, the same programming logic is used to generate the indices in both directions. The only difference between the two is that the table columns of the source and target languages are swapped when provided to the program for the Slavonic-Greek index and for the Greek-Slavonic index, respectively. For shortness, in the next section we will only show the two-way result in one of the dictionaries – the one that better illustrates the differences.

## 4. Solutions

In order to illustrate in detail how our approach works, here we look at specific problems and our corresponding solutions. We address two categories of problems: 1) quantitative asymmetry, which we illustrate with the generated lists, and 2) variability in transcripts, better illustrated by the resulting indices.

### 4.1. Quantitative asymmetry

As mentioned in the previous section, with our approach and the program implementing it, the quantitative asymmetry is solved at the *adaptation* step. Several examples that demonstrate different cases of asymmetry follow.

|   | E | F | G | H | I | J | K | L | M |
|---|---|---|---|---|---|---|---|---|---|
|   | 10/59a12 | ꙁакона | и єгда ꙁакона хо- | ꙁаконъ | ꙁаконъ дати |   |   | νομοθετεῖν | **νομοθετέω** |
|   | 10/59a13 | дати | тѣаше дати нюе- | дати |   |   |   |   |   |

дати
| ꙁаконъ дати
    νομοθετέω: ꙁакона дати/νομοθετεῖν (10/59a12-13)
ꙁаконъ
| ꙁаконъ дати
    νομοθετέω: ꙁакона дати/νομοθετεῖν (10/59a12-13)

νομοθετέω
    ꙁаконъ дати → ꙁаконъ & дати: νομοθετεῖν/ꙁакона
    дати (10/59a12-13)

**Figure 4:** An example input (above) and output (below) for translation asymmetry one-to-many (1:n).

**One-to-many**    (1:n) is the base example that occurs when a Greek word is translated with a Slavonic phrase. To indicate such occurrences, we use background colouring of the corresponding rows in the word-use column (F), as shown in Fig. 4. In cases where two groups immediately follow each other, they are delineated by using an empty line between the two.

What the *adaptation* step does is to collect translation information from the whole phrase and associate it with each of the lexemes by adding it in its corresponding row in the table. In the example below, according to the currently generated dictionary, the target words (for Slavonic in column F) remain as they are, but the lemmas (H) are given as a phrase – together – and repeated for each of resulting entries.

**One-to-many, including grammatical value**    (1:n*) is a more complicated case where one of the words from the phrase has only grammatical meaning. The example in Fig. 5 uses an analytic verb form for a past participle.

To indicate such occurances we use "gramm." in the second lemma (column I) with a coloured background. The particularity in this case is that the grammatical value is valid only for the lexeme on the same line, but not for the rest of the phrase. In the dictionaries, therefore, this information should be shown only for the corresponding lemma, but not for the others in the group.

In this case, the *adaptation* step does not add the sublemmas of the lexeme with grammatical value to the information about the other lexemes. Thus, the corresponding feature is present

|   | E | F | G | H | I | J | K | L | M |
|---|---|---|---|---|---|---|---|---|---|
|   | 1/6b16 | речено | о о͠цн речено єстъ. | рещи |   |   |   | εἴρηται | **λέγω** |
|   | 1/6b16 | єстъ | о о͠цн речено єстъ. | быти | gramm. |   |   |   | pass. |

быти
|| gramm.
    pass.: речено єстъ/εἴρηται (1/6b16)
рещи
    λέγω: речено єстъ/εἴρηται (1/6b16)

λέγω
    рещи: εἴρηται/речено єстъ (1/6b16)
pass.
    быти gramm.: εἴρηται/речено єстъ (1/6b16)

**Figure 5:** Input and output for translation asymmetry one-to-many with grammatical value (1:n*).

| E | F | G | H | I | J | K | L | M | N |
|---|---|---|---|---|---|---|---|---|---|
| 04/18a11 | доброу | ша са· и доброу | добръ | добрѫ жнzнь нмѣтн | | | ἀπολαύοντας εὐημερίας | ἀπολαύω | ἀπολαύω εὐημερίας |
| 04/18a12 | жнzнь | жнzнь нмоу- | жнzнь | | | | εὐημερίας | εὐημερία | |
| 04/18a12 | нмоуща· | жнzнь нмоу- | нмѣтн | | | | | | |

добръ
| добрѫ жнzнь нмѣтн
    ἀπολαύω εὐημερίας → ἀπολαύω & εὐημερία: доброу жнzнь
    нмоуща·/ἀπολαύοντας εὐημερίας (4/18a11-12)
жнzнь
| добрѫ жнzнь нмѣтн
    ἀπολαύω εὐημερίας → ἀπολαύω & εὐημερία: доброу жнzнь
    нмоуща·/ἀπολαύοντας εὐημερίας (4/18a11-12)
нмѣтн
| добрѫ жнzнь нмѣтн
    ἀπολαύω εὐημερίας → ἀπολαύω & εὐημερία: доброу жнzнь
    нмоуща·/ἀπολαύοντας εὐημερίας (4/18a11-12)

ἀπολαύω
| ἀπολαύω εὐημερίας
    добрѫ жнzнь нмѣтн → добръ & жнzнь & нмѣтн: ἀπολαύοντας
    εὐημερίας/доброу жнzнь нмоуща· (4/18a11-12)
εὐημερία
| ἀπολαύω εὐημερίας
    добрѫ жнzнь нмѣтн → добръ & жнzнь & нмѣтн: ἀπολαύοντας
    εὐημερίας/доброу жнzнь нмоуща· (4/18a11-12)

**Figure 6:** An example input and output for many-to-many asymmetry (n:m).

only in the translation from Slavonic to Greek and leads to an asymmetry between the two dictionaries in Fig. 6 (see sublemma "gramm."). This is the reason why different adapted tables are needed for the aggregation step from Slavonic to Greek and from Greek to Slavonic.

**Many-to-many** (n:m) asymmetry is an occurrence when a Greek phrase is translated with a Slavonic phrase but there is no direct corresponding between the separate words of these phrases in the two languages as shown in Fig. 6. Our approach solves this case using the same logic. In the adaptation step, the program collects in a single row all the information needed for an entry in the dictionaries. As a consequence, the *aggregation* step, does not need to perform any processing other than just reading the data row by row and adding it to the *sorted mapping* data structure.

### 4.2. Variant readings in manuscript copies

As a rule, the indices and lists need to include all word uses from the main text and those variants from the other copies that represent either better readings than the main text or equally possible readings in the given context. Missing, wrong or less precise variants from the Slavonic transcripts W, G and H are ignored, and not lemmatised, so the program does not include them in the dictionaries.

**Different translation correspondences** (seen in Fig. 1 and Fig. 2) is a scenario where a word other than the main one is used in the additional sources. In such circumstances, the dictionaries show the usage in both lexemes used (in the main version and in the variant), adding references to the variant lexemes to the addresses of the usages. This functionality is the reason why – in the *aggregation* step – the structure describing the usages also contains information about the alternatives.

| A | B | C | D | E | F | G | H | I | J | K | L | M |
|---|---|---|---|---|---|---|---|---|---|---|---|---|
| ѥдьнородоу H | ѥдьнородъ | | | 1/W168a34 | ѥдьноѥдоу | въ имѣти ѥдьноѥдоу | ѥдьночѧдъ | | | | μονογενῆ | μονογενής |
| | | | | 1/W168a28 | ѥдьноѥдаго | йже ѥдьноѥдаго ѹтрокъ | ѥдьночѧдъ | | | | μονογενοῦς | μονογενής |
| ннѹѥдаго G  ѥдьнородоу H | ѥдьнородъ H / инѹчѧдъ G | | | 1/W168a25 | ѥдьноѥдоу | кргы(!) гле· сладкѹ ıако | ѥдьночѧдъ | | | | *μονογενοῦς* | *μονογενής* |
| ѥдьноѥды WH Ø G | ѥдьночѧдъ | | | 1/5a4 | инѹчѧдын | нъ ıако бъ· а инѹчѧдн | ннѹчѧдъ | | | | μονογενής | μονογενής |

**ннѹчѧдъ** (1 + 1ᵛᵃʳ)
- μονογενής (2): 1/5a4 » [ѥдьночѧдъᵂᴴ]; *1/W168a25*ᴳ » ѥдьночѧдъ W [ѥдьнородъᴴ]

**ѥдьнородъ** (2ᵛᵃʳ)
- μονογενής (2): *1/W168a25*ᴴ » ѥдьночѧдъ W [инѹчѧдъᴳ]; 1/W168a34ᵂᴴ » ѥдьночѧдъ W

**ѥдьночѧдъ** (3 + 1ᵛᵃʳ)
- μονογενής (4): 1/5a4ᵂᴴ » ннѹчѧдъ S; *1/W168a25* » [ѥдьнородъᴴ, инѹчѧдъᴳ]; 1/W168a28; 1/W168a34 » [ѥдьнородъᴴ]

**μονογενής** (4)
- ннѹчѧдъ (1 + 1ᵛᵃʳ): 1/5a4 » [ѥдьночѧдъᵂᴴ]; *1/W168a25*ᴳ » ѥдьночѧдъ W [ѥдьнородъᴴ]
- ѥдьнородъ (2ᵛᵃʳ): *1/W168a25*ᴴ » ѥдьночѧдъ W [инѹчѧдъᴳ]; 1/W168a34ᴴ » ѥдьночѧдъ W
- ѥдьночѧдъ (3 + 1ᵛᵃʳ): 1/5a4ᵂᴴ » ннѹчѧдъ S; *1/W168a25* » [ѥдьнородъᴴ, инѹчѧдъᴳ]; 1/W168a28; 1/W168a34 » [ѥдьнородъᴴ]

**Figure 7:** An example of multiple correspondences from different copies. Note: First three rows use non-standard addresses, because they are from a snippet missing in the main copy (S) and published according to ÖNB 12 (W).

**Different copies suggest different correspondences** is a fundamentally similar but more complicated occurrence. In the example in Fig. 7, the rows indicate four different combinations delineating three different translations (ннѹчѧдъ, ѥдьнородъ and ѥдьночѧдъ) of the Greek word μονογενής, as well as the possibility of a missing translation. Similar to the previous example, this diversity should be traceable in the generated lists and indices through the corresponding references. Variant readings in copies are relatively rare, and it is even rarer to have two different variant readings to the same lexeme. The solution should therefore be able to handle such cases without complicating the manual enrichment and annotation performed by philologists. The solution adopted here combines the different variant readings in column A and their respective lemmata in column B for the Slavonic variant. Such cases are interpreted in the *aggregation* step, where the combinations of variants are recorded as separate *Alignment* data objects.

## 5. Conclusion

This paper presents our work on creating dictionaries for UE. In it, we show the combined solution to two particular methodological challenges: quantitative asymmetry and variation in the sources. We solve these problems applying a stepwise approach of adaptation, aggregation and export. We demonstrate a *divide-and-conquer* approach (implemented in the *adaptation* step) that allows solving even complex cases of *many-to-many* quantitative asymmetry with different grammatical values. We also propose alignment modelling for the purpose of generating dictionaries with cross-references for text variants. As a result, in the dictionaries it is possible to indicate not only the alignment between original and translation, but also between these two and a large number of variants.

The approach has been applied to the 52 constituent texts of the UE (the introduction and the 51 sermons). More complex combinations of asymmetry and variation in the sources were also successfully modelled for this purpose. Such cases represent combinations of the examples provided in this paper. The program was written with applicability to other Greek-Slavonic translations in mind. In order to allow for this applicability, and for future extensions, we commit to publishing the software code under a free license.

## Acknowledgments

The article is written within the project *The Vocabulary of Constantine of Preslav's Uchitel'noe evangelie ('Didactic Gospel'): Old Bulgarian-Greek and Greek-Old Bulgarian Word Indices*, financed by the Bulgarian National Science Fund (contract КП-06-Н50/2 of 30.11.2020).